\documentclass[letterpaper, 10 pt, conference]{ieeeconf}  

\IEEEoverridecommandlockouts                              

\usepackage[backend=bibtex,
            hyperref=false,
            url=false,
            isbn=false,
            doi=false,
            backref=false,
            style=numeric-comp,
            sortcites,
            block=none]{biblatex}
            
\renewcommand{\bibfont}{\small}
\usepackage{multirow}

\usepackage{algpseudocode}
\usepackage{algorithmicx}
\usepackage[ruled]{algorithm}

\usepackage{color,xcolor}
\usepackage{epsfig}
\usepackage{graphicx}

\usepackage{adjustbox}
\usepackage{array}
\usepackage{booktabs}
\usepackage{colortbl}
\usepackage{float,wrapfig}
\usepackage{hhline}
\usepackage{multirow}
\usepackage{subcaption} 

\usepackage{bm}
\usepackage{nicefrac}
\usepackage{microtype}

\usepackage{changepage}
\usepackage{extramarks}
\usepackage{fancyhdr}
\usepackage{lastpage}
\usepackage{setspace}
\usepackage{soul}
\usepackage{xspace}

\usepackage{url}

\usepackage{enumerate}

\usepackage{titlesec}
\usepackage{makecell}
\usepackage{mathtools}
\usepackage{stmaryrd}
\usepackage{hyperref}
\hypersetup{
    colorlinks=true,
    linkcolor=blue,
    filecolor=magenta,      
    urlcolor=cyan,
    }

\newcolumntype{L}[1]{>{\raggedright\let\newline\\\arraybackslash\hspace{0pt}}m{#1}}
\newcolumntype{C}[1]{>{\centering\let\newline\\\arraybackslash\hspace{0pt}}m{#1}}
\newcolumntype{R}[1]{>{\raggedleft\let\newline\\\arraybackslash\hspace{0pt}}m{#1}}

\makeatletter
\DeclareRobustCommand\onedot{\futurelet\@let@token\@onedot}
\def\@onedot{\ifx\@let@token.\else.\null\fi\xspace}

\g@addto@macro\normalsize{%
  \setlength\abovedisplayskip{0pt}
  \setlength\belowdisplayskip{0pt}
  \setlength\abovedisplayshortskip{0pt}
  \setlength\belowdisplayshortskip{0pt}
}
\makeatother

\newcommand{\ours}{Neural Descriptor Field\xspace}
\newcommand{\ourshort}{NDF\xspace}
\newcommand{\don}{DON\xspace}

\newcommand{\sect}[1]{Section~\ref{#1}}

\newcommand{\ignore}[1]{}

\newcommand{\coord}{\mathbf{x}}

\newcommand{\SE}[1]{\text{SE}(#1)}

\newcommand{\sethree}{$\text{SE}(3)$}

\newcommand{\encoder}{\mathcal{E}}
\newcommand{\sothree}{$\text{SO}(3)$}

\newcommand{\pointcloud}{\mathbf{P}}
\newcommand{\object}{\mathbf{S}}
\newcommand{\pose}{\mathbf{T}}
\newcommand{\pointndf}{f}
\newcommand{\posendf}{F}
\newcommand{\pointdescriptor}{z}

\newcommand{\posedescriptor}{\mathcal{Z}}
\newcommand{\demopoint}{\hat{\textbf{x}}}
\newcommand{\demopointcloud}{\hat{\pointcloud}}
\newcommand{\democoord}{\hat{\coord}}
\newcommand{\demopose}{\hat{\pose}}

\newcommand{\demoposerel}{\pose_{\text{rel}}}
\newcommand{\descriptor}{\mathcal{Z}}

\newcommand{\probes}{\mathcal{X}}

\usepackage{color}
\definecolor{turquoise}{cmyk}{0.65,0,0.1,0.3}
\definecolor{purple}{rgb}{0.65,0,0.65}
\definecolor{dark_green}{rgb}{0, 0.5, 0}
\definecolor{orange}{rgb}{0.8, 0.6, 0.2}
\definecolor{red}{rgb}{0.8, 0.2, 0.2}
\definecolor{darkred}{rgb}{0.6, 0.1, 0.05}
\definecolor{blueish}{rgb}{0.0, 0.3, .6}
\definecolor{light_gray}{rgb}{0.7, 0.7, .7}
\definecolor{pink}{rgb}{1, 0, 1}
\definecolor{greyblue}{rgb}{0.25, 0.25, 1}

\definecolor{blueish}{rgb}{0.0, 0.3, .6}

\DeclareRobustCommand{\vsnote}[1]{\ifthenelse{\boolean{draft-mode}}{\textcolor{dark_green}{[VS: #1}]}{}}

\DeclareRobustCommand{\pa}[1]{\ifthenelse{\boolean{draft-mode}}{\textcolor{green}{\textbf{[PA: #1}]}}{}}
\DeclareRobustCommand{\asnote}[1]{\ifthenelse{\boolean{draft-mode}}{\textcolor{blue}{\textbf{AS: #1}}}{}}
\DeclareRobustCommand{\arnote}[1]{\ifthenelse{\boolean{draft-mode}}{\textcolor{green}{\textbf{AR: #1}}}{}}

\newcommand{\Fig}[1]{Fig.~\ref{fig:#1}}
\newcommand{\Figure}[1]{Fig.~\ref{fig:#1}}

\newcommand{\Table}[1]{Table~\ref{tbl:#1}}
\newcommand{\eq}[1]{(\ref{eq:#1})}

\usepackage{blindtext}

\renewcommand{\paragraph}[1]{\vspace{.1em}\noindent\textbf{#1}.}

\usepackage{amsmath,amsfonts,bm}

\def\eqref#1{equation~\ref{#1}}

\def\1{\bm{1}}

\DeclareMathAlphabet{\mathsfit}{\encodingdefault}{\sfdefault}{m}{sl}
\SetMathAlphabet{\mathsfit}{bold}{\encodingdefault}{\sfdefault}{bx}{n}

\DeclarePairedDelimiterX{\infdivx}[2]{(}{)}{%
  #1\;\delimsize|\delimsize|\;#2%
}

\title{\ours{}s:\\ \sethree-Equivariant Object Representations for Manipulation}

\author{
\authorblockA{Anthony Simeonov$^{*,1}$, Yilun Du$^{*,1}$, Andrea Tagliasacchi$^{2,3}$, \\Joshua B. Tenenbaum$^1$, Alberto Rodriguez$^1$, Pulkit Agrawal$^{\dagger, 1}$, Vincent Sitzmann$^{\dagger, 1}$ \\ 
$^1$Massachusetts Institute of Technology $\quad$ $^2$Google Research $\quad$ $^3$University of Toronto\\
$^*$Authors contributed equally, order determined by coin flip. $^\dagger$Equal Advising. \\}
}

\newboolean{draft-mode}
\setboolean{draft-mode}{true}

\begin{document}

\twocolumn[{%
\renewcommand\twocolumn[1][]{#1}%
\maketitle
\begin{center}
\centering
\captionsetup{type=figure}
\includegraphics[width=\linewidth]{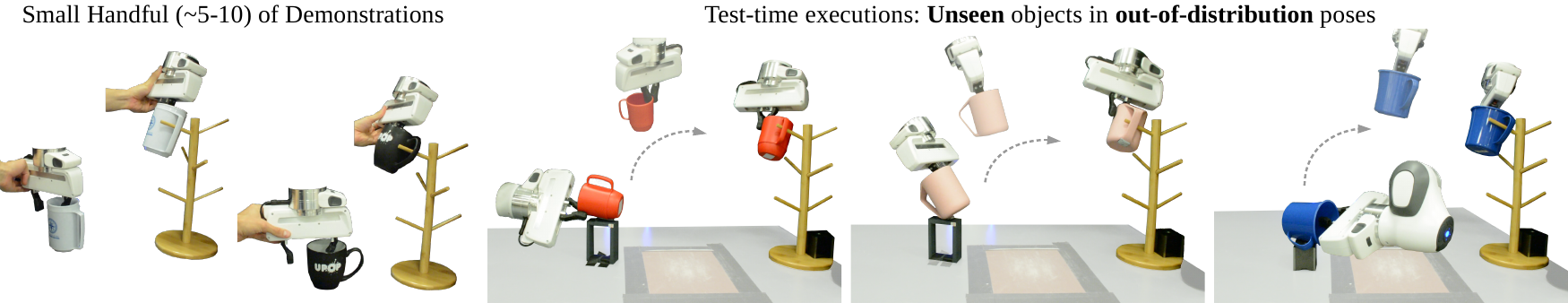}
\captionof{figure}{\small{Given a few ($\sim$5-10) demonstrations of a manipulation task (left), \ours{}s (\ourshort{}s) generalize the task to novel object instances in any 6-DoF configuration, \emph{including those unobserved at training time}, such as mugs with arbitrary 3D translation and rotation (right).
\ourshort{}s are continuous functions that map 3D spatial coordinates to spatial descriptors. We generalize this to functions which encode \sethree{} poses, such as those used for grasping and placing.
\ourshort{}s are trained self-supervised for the surrogate task of 3D reconstruction, do not require labeled keypoints, and are \sethree{}-equivariant, guaranteeing generalization to unseen object configurations.}}

\label{fig:teaser}
\vspace{-5pt}
\end{center}

}]

\global\csname @topnum\endcsname 0
\global\csname @botnum\endcsname 0

\thispagestyle{empty}
\pagestyle{empty}

\begin{abstract}
We present \ours{}s (\ourshort{}s), an object representation that encodes both points and relative poses between an object and a target (such as a robot gripper or a rack used for hanging) via category-level descriptors.
We employ this representation for object manipulation, where given a task demonstration, we want to repeat the same task on a \textit{new} object instance from the \textit{same} category.
We propose to achieve this objective by searching (via optimization) for the pose whose descriptor matches that observed in the demonstration.
\ourshort{}s are conveniently trained in a self-supervised fashion via a 3D auto-encoding task that does not rely on expert-labeled keypoints.
Further, \ourshort{}s are \sethree{}-equivariant, guaranteeing performance that generalizes across all possible 3D object translations \textit{and} rotations.
We demonstrate learning of manipulation tasks from few ($\sim$5-10) demonstrations both in simulation and on a real robot.
Our performance generalizes across both object instances and 6-DoF object poses, and significantly outperforms a recent baseline that relies on 2D descriptors. Project website: \href{https://yilundu.github.io/ndf/}{https://yilundu.github.io/ndf/}


\end{abstract}


\section{Introduction}
\label{sec:intro}
Task demonstrations are an intuitive and a powerful mechanism for communicating complex tasks to a robot~\cite{argall2009survey,pomerleau1989alvinn,schaal1999imitation}. However, the ability of current methods to learn from demonstrations is severely limited. Consider the task of teaching the robot to pick up a mug and place it on a rack. After learning, if we want the robot to place a novel instance of a mug from any starting location and orientation, state-of-the-art systems would require a large number of demonstrations spanning the space of different initial positions, orientations and mug instances. This requirement makes it extremely tedious to communicate tasks using demonstrations. Moreover, this approach based on data augmentation comes with no algorithmic guarantees to generalization to out-of-distribution object configurations.

Our goal is to build a robotic system that can learn such pick-and-place tasks for unseen objects in a data-efficient manner.
%
In particular, we desire to construct a system which can manipulate objects from the same category into target configurations, irrespective of the object's 3D location and orientation (see Figure~\ref{fig:teaser}) from just a few training demonstrations ($\sim 5-10$).


Consider the task of picking a mug. When task and demonstration objects are identical, the robot can pick up the object by transferring the demonstrated grasp to the new object configuration. For this it suffices to attach a coordinate frame to the demonstration mug, estimate the pose of this frame on the new mug, and move the robot to the relative grasp pose that was recorded in the demonstration with respect to the coordinate frame.
%
%
Let us now consider mugs that vary in shape and size, wherein grasping requires aligning the gripper to a \emph{local} geometric feature whose location \emph{varies} depending on the shape of the mug. In this case, estimating the coordinate frame on the new mug and moving to the relative grasp pose recorded in the demonstration will fail, \emph{unless} the frame is attached to the specific geometric feature that is used for grasping. However, the choice of which geometric feature to use is \emph{under-specified} unless we consider the task, and different tasks require alignment to \emph{different} features. 

%
For example, to imitate \emph{grasping} along the rim, we may require to define a local frame such that identical gripper poses expressed in this frame all lead to grasping along the rim, irrespective of the height of the mug. 
%
On the other hand, imitating a demonstration of object \emph{placing} may require a \emph{new} coordinate frame that can align a placing target (e.g., a shelf) to the bottom surface of the mugs. 
These examples elucidate that the relevant geometric structure for alignment is \emph{task-specific}.
Having identified the task-relevant geometric feature, we can attach a coordinate frame and measure the pose of the gripper/placing target in task demonstrations relative to this feature on different object instances. Given a new object instance, the task can be performed by first identifying the local coordinate frame on the object and then obtaining a relative grasping/placing pose in this coordinate frame that is consistent with the demonstrations. 

The two key questions to be answered in this process are: (1) how to specify the relevant feature and local frame for a given task; and (2) how to solve for the corresponding frame given a new object instance.
%
Prior approaches address these questions by hand-labeling a large dataset of task-specific keypoints and training a neural network to predict their location on new instances~\cite{manuelli2019kpam, gao2021kpam}. Detected keypoints are then used to recover a local coordinate frame. However, collecting this dataset for each task is expensive, and these methods fail to generalize to new instances in the regime of few demonstrations.
To mitigate the generalization issue, other prior work first learns to model \emph{dense} point correspondence in a task-agnostic fashion~\cite{florence2018dense}. At test time, human-annotated keypoints are individually and independently corresponded one-by-one, and the local coordinate frame is established via registration to the keypoints on the demonstration instance.
%
This enables imitation from few demonstrations, but current approaches---which operate in 2D---suffer several key limitations.
(i) Keypoints may only lie on the surface of the object, making it difficult to encode important free-space locations (i.e., in the center of a handle). Further, if the object is partially occluded, keypoint locations cannot be inferred.
(ii) Small errors in estimating the corresponding location of each keypoint can result in large errors in solving for the transform and consequently the resulting coordinate frame. 
(iii) Existing methods are not equivariant to \sethree{} transformations and thus not guaranteed to provide correct correspondence when instances are in unseen poses. 
(iv) Human keypoint annotation is required to identify task-specific features.

We propose a novel method to encode dense correspondence across object instances, dubbed \ours{}s (\ourshort{}), that effectively overcomes the limitations of prior work:
(i) We represent an object point cloud $\pointcloud$ as a continuous function $\pointndf(\coord|\pointcloud)$
that maps any 3D coordinate $\coord$ to a spatial descriptor. 
Descriptors encode the spatial relationship of $\coord$ to the salient geometric features of the object in a way that is consistent across different shapes in a category of objects. 
Coordinates are not constrained to be on the object and can potentially be occluded. 
%
%

(ii) We represent a coordinate frames associated with a local geometric structure using a \emph{rigid set} of query points. The configuration of these points is represented as an \sethree{} pose with respect to a canonical pose in the world frame. For each object instance in the demonstration, the query point set is converted into a set of feature descriptors by concatenating point descriptors of all the points. The feature representation resulting from evaluating query points at different \sethree{} transformations forms what we call a \emph{pose descriptor field}. To estimate the pose of the local coordinate frame on a new object instance, we optimize for the \sethree{} transformation of the query points that minimizes the distance of feature descriptors with those of the demonstration objects. This process solves for feature matching and the coordinate frame's pose jointly, instead of the two-step process employed by prior work which is more prone to errors. Furthermore, because query points are defined in 3D (as opposed to 2D keypoints) and it is not necessary that location of all query points is observed, the proposed procedure for finding coordinate frame is more robust and has higher accuracy than existing methods.    

(iii) To guarantee that we can successfully estimate the local frame for all 6-DoF configurations of the test-time object instance (i.e., generalization), we construct the pose descriptor fields to be  equivariant to $\sethree{}$ transformations. For this we leverage recent progress in geometric deep learning~\cite{deng2021vector}. 
%
(iv) Finally, we devise a procedure for using the demonstrations to obtain the set of query points, such that the pose descriptor - and thus, the recovered coordinate frame - is sensitive to task-relevant local geometric features, overcoming the need for human-annotated keypoints.

Using this novel formulation, we propose a system that can imitate pick-and-place tasks for a category of objects from only a small handful of demonstrations. On three unique pick-and-place tasks, \ours{}s enables both pick and place of unseen object instances in out-of-distribution configurations with an overall success rate above 85\%, using only 10 expert demonstrations and consistently outperforms baselines that operate in 2D and are not \sethree{} equivariant. 

\begin{figure*}[t]
\begin{center}
\begin{minipage}{\columnwidth}
\includegraphics[width=\columnwidth]{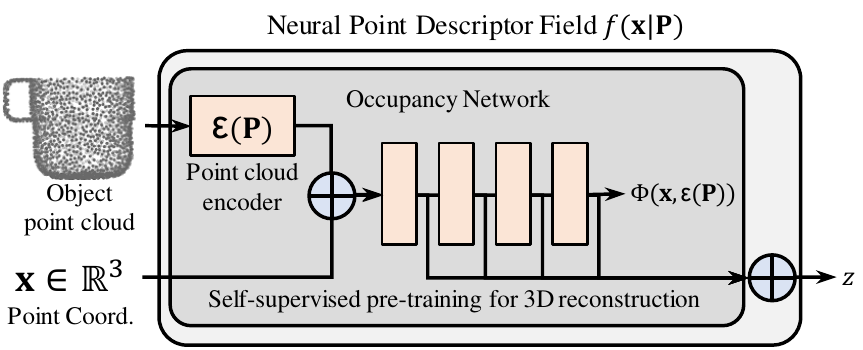}
\caption{\small \textbf{Point Descriptor Fields} -- We propose to parameterize a Neural Point Descriptor Field $f$ as the concatenation of the layer-wise activations of an occupancy network $\Phi(\coord, \encoder(\pointcloud))$.
Both the point cloud encoder and the point descriptor function can be pre-trained with a 3D reconstruction task.}
\label{fig:overview-left}
\end{minipage}
\hfill
\begin{minipage}{\columnwidth}
\includegraphics[width=\columnwidth]{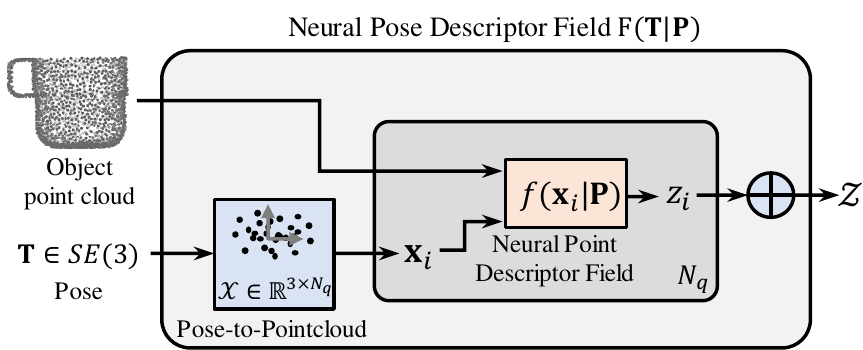}
\caption{\small \textbf{Pose Descriptor Fields} -- \ourshort{}s can extract \emph{pose} descriptors by representing a pose via its action on a \emph{query pointcloud} $\probes$, and then extracting point-level spatial descriptors $z_i$ for each point $\coord_i$ with a point-level \ourshort{}. Concatenation then yields the final pose descriptor $\descriptor$.}
\label{fig:overview-right}
\end{minipage}
\end{center}
\vspace{-25pt}
\end{figure*}

\section{Method}
We present a novel representation that models dense correspondence across object instances at the level of points and local coordinate frames. 
Our representation enables an intuitive mechanism for specifying a task-relevant local frame using a demonstration task and point cloud $\demopointcloud$, 
along with the efficient and robust computation of a corresponding local frame when presented with a new point cloud $\pointcloud$.
%

In \sect{sect:pointndfs}, we introduce a continuous function  $\pointndf(\coord | \pointcloud)$  that maps a 3D coordinate $\coord$ and a point cloud $\pointcloud$ to a spatial descriptor that encodes information about the spatial relationship of $\coord$ to the category-level geometric features of the object.
%
We demonstrate that we can represent this function using a neural network trained in a task-agnostic manner via 3D reconstruction, and that this training objective learns descriptors that encode point-wise correspondence across a category of shapes. 
%
We furthermore show how we may equip these point descriptor fields with \sethree{}-equivariance, enabling correspondence matching across object instances in arbitrary \sethree{} poses.
%
In \sect{sect:posendfs}, we leverage these point descriptors to establish correspondence for a rigid \emph{set} of points, whose configuration is used to parameterize a local coordinate frame near the object. This enables us to \emph{directly solve for the \sethree{} pose} of the transformed point set whose descriptors best match a reference descriptor set, and recover the corresponding local frame relative to a new object. 

%
%
%
We then discuss how to apply this novel representation for transferring grasp and place poses from a set of pick-and-place demonstrations: We first show how contact interactions between the manipulated object and known external rigid bodies (such as a gripper, rack, or shelf) can be used to sample query points near important geometric features, yielding descriptors for \emph{task-relevant} local reference frames directly from demonstrations. 
Finally, in \sect{sect:obj_manip}, we show how we use pose descriptor fields and a small handful of demonstrations to reproduce a pick-and-place task on a new object in an arbitrary initial pose. 

\subsection{Neural Point Descriptor Fields}
\label{sect:pointndfs}
%
Our key idea is to represent an object as a function $\pointndf$ that maps a 3D coordinate $\coord$ to a spatial descriptor $\pointdescriptor=\pointndf(\coord)$ of that 3D coordinate:
\begin{align}
    \pointndf(\coord): \mathbb{R}^3 \to \mathbb{R}^n 
\end{align}
$\pointndf$ may further be conditioned on an object point cloud $\pointcloud \in \mathbb{R}^{3\times N}$ to output \emph{category-level} descriptors $\pointndf(\coord | \pointcloud)$. 
%
We propose to parameterize $\pointndf$ via a neural network. This yields a \emph{differentiable} object representation that continuously maps \emph{every} 3D coordinate to a spatial descriptor. As we will see, this continuous, differentiable formulation enables us to find correspondence across object instances via simple first-order optimization.
Finally, it remains to learn the weights of a neural descriptor field. On first glance, this would require setting up a training objective for correspondence matching, and consequently, collection and labeling of a custom dataset. Instead, we propose and demonstrate that we may leverage recently proposed neural implicit shape representations~\cite{mescheder2019occupancy,park2019deepsdf,chen2019learning} to parameterize $\pointndf$ and learn its weights in a self-supervised manner.
%
%

\paragraph{Background: neural implicits}
Neural implicit representations represent the 3D surface of a shape as the level-set of a neural network.
In particular, Mescheder et al.~\cite{mescheder2019occupancy} represent a 3D shape as an MLP $\Phi$ that maps a 3D coordinate $\coord$ to its occupancy value:
\begin{equation}
    \Phi(\coord): \mathbb{R}^3 \to [0,1]
\end{equation}
We are interested in learning a low-dimensional latent space of 3D shapes, which can be achieved by parameterizing the latent space with a latent code $\mathbf{v} \in \mathbb{R}^k$ and concatenating it with $\coord$, encoding different shapes via different latent codes. 
These latent codes are obtained as the output of a PointNet~\cite{qi2017pointnet++}-based point cloud encoder $\encoder$ that takes as input a point cloud~$\pointcloud$, leading to a conditional occupancy function:
\begin{gather}
    \Phi(\coord, \encoder(\pointcloud)): \mathbb{R}^3 \times \mathbb{R}^k \to \left[0,1\right]
    \label{eq:occupancy}
\end{gather}
The full model can be trained end-to-end on a dataset of partial point clouds and corresponding occupancy voxelgrids of the objects' full 3D geometry, thus learning to predict the occupancy of a complete 3D object from a \textit{partial} pointcloud. This is an attractive property, as at test time, we regularly only observe partial point clouds of objects due to occlusions.

\paragraph{Neural feature extraction -- \Fig{overview-left}}
To enable category-level object manipulation, a spatial descriptor for a coordinate $\coord$ given a point cloud $\pointcloud$ should encode information about the spatial relationship of $\coord$ to the salient features of the object. That is, for mugs, descriptors should encode information about how far $\coord$ is away from the mug's handle, rim, etc. 

Our key insight is that the category-level 3D reconstruction objective trains $\Phi(\coord, \encoder(\pointcloud))$ to be a hierarchical, coarse-to-fine feature extractor that encodes exactly this information:
$\Phi$ is a classifier whose decision boundary is the surface of the object. Intuitively, each layer of $\Phi$ is a set of ReLU hyperplanes that are trained to encode \emph{how far a given coordinate $\coord$ is away from this decision boundary}, such that ultimately, the final layer may classify it as \emph{inside} or \emph{outside} the shape, where layers encode increasingly finer surface detail.
The output of the pointcloud encoder $\encoder(\pointcloud)$ in turn determines \emph{where} this decision boundary lies in terms of a small set of latent variables. 
This bottleneck forces the model to use these few latent variables to parameterize the salient features of the object category, which is impressively demonstrated by smooth latent-space interpolations and unconditional shape samples ~\cite{park2019deepsdf,mescheder2019occupancy, chen2021learning}. Prior work has leveraged this property of the activations of $\Phi$ to classify which semantic part of an object a given coordinate $\coord$ belongs to~\cite{kohli2020semantic}, a task which is closely related to modeling correspondence across a category. 

We thus propose to parameterize our neural point descriptor field $\pointndf(\coord | \pointcloud)$ as the function that maps every 3D coordinate $\coord$ to the \emph{vector of concatenated activations} of $\Phi$: 
%
\begin{align}
    \pointndf(\coord | \pointcloud) = \bigoplus_{i=1}^L \Phi^i(\coord, \encoder(\pointcloud))
    \label{eq:pointndf}
\end{align}
with the activation of the $i$th layer as $\Phi^i$, total number of layers $L$, and concatenation operator $\bigoplus$. We choose to concatenate activations across layers to encourage consideration of features across scales and ablate this effect in \Table{ablation}.

\paragraph{Equivariance w.r.t. \sethree}
%
%
A key requirement of our descriptor field is to ensure descriptors remain constant if the position of $\coord$ relative to $\pointcloud$ remains constant, regardless of their global configuration in the world coordinate system.
In other words, we require $\pointndf$ to be \emph{invariant} to joint transformation of $\coord$ and $\pointcloud$, implying the descriptor field should be \emph{equivariant} to \sethree{} transformations of $\pointcloud$ --
%
we wish that if an object is subject to a rigid body transform $(\mathbf{R}, \mathbf{t}) \in$ \sethree{} its spatial descriptors transform accordingly:
\begin{equation}
    \pointndf(\coord | \pointcloud) \equiv \pointndf(\mathbf{R} \coord + \mathbf{t} | \mathbf{R} \pointcloud + \mathbf{t}).
\end{equation}
Translation equivariance is conveniently implemented by subtracting the center of mass of the point cloud from both the input point cloud and the input coordinate. We thus re-define $\pointndf(\coord|\pointcloud)$ as:
\begin{equation}
    \pointndf(\coord | \pointcloud) = \pointndf(\coord - \mathbf{\mu} | \pointcloud - \mathbf{\mu}); \quad \mu = \frac{1}{N} \sum_{i=1}^N  \pointcloud_i
\end{equation}
This results in the input to $\pointndf$ always being zero-centered, irrespective of the absolute position of $\pointcloud$, making $\pointndf$ invariant to joint translations of $\coord$ and $\pointcloud$.
To achieve \textit{rotation} equivariance, we rely on recently proposed Vector Neurons~\cite{deng2021vector}, which propose a network architecture that equips an occupancy network, i.e., the composition of $\encoder$ and $\Phi$ in \eq{occupancy}, with full \sothree{} equivariance. 
By replacing $\Phi(\coord,\encoder(\pointcloud))$ in ~\eq{pointndf} with this \sothree{}-equivariant architecture,  $\pointndf$ immediately inherits this property, such that for $\mathbf{R} \in$ \sothree{}:
\begin{equation}
    \pointndf(\coord|\pointcloud) \equiv \pointndf(\mathbf{R} \coord | \mathbf{R} \pointcloud)
\end{equation}
Combining this with the pointcloud mean-centering scheme yields complete \sethree{} equivariance --- i.e., $\pointndf$ now enjoys a \emph{guarantee} that transforming an input pointcloud by \emph{any} \sethree{} transform will transform the locations of spatial descriptors accordingly, leaving them unchanged otherwise.
This guarantees that we can generalize to arbitrary object poses, including those completely unobserved at training time.

\begin{figure}[t]
\includegraphics[width=\columnwidth]{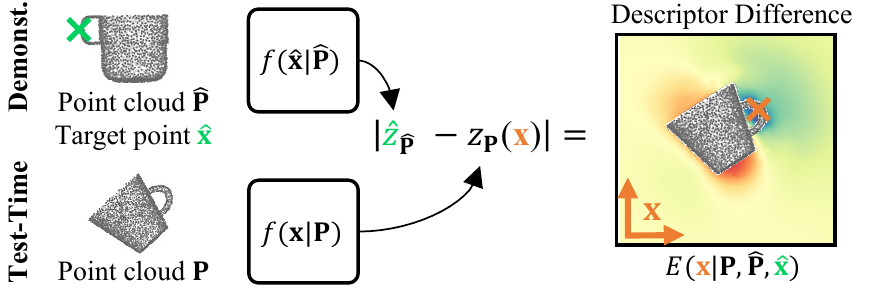}
\caption{\small \textbf{Energy landscape induced by \ourshort{}s} --
Given a demonstration in the form of a pointcloud-point tuple $(\demopointcloud,\democoord)$, and the pointcloud of an unseen object instance $\pointcloud$, \ourshort{}s induce an energy landscape whose minimizer is the equivalent point for the unseen object. This energy is differentiable w.r.t. the point coordinates.
}
\label{fig:energy-left}
\vspace{-20pt}
\end{figure}

\begin{figure}[t]
\centering
\includegraphics[width=0.95\columnwidth]{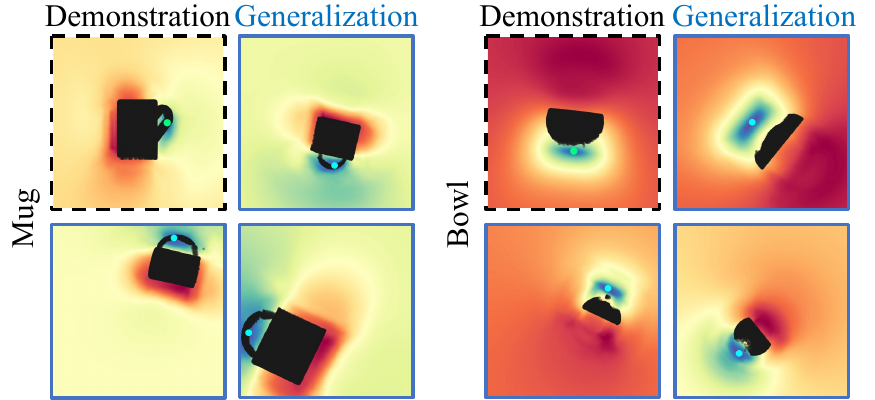}
\caption{\small
\textbf{Equivariance and generalization of \ourshort{}s} --
Absolute descriptor differences for a 2D target point $\democoord \in \mathbb{R}^2$.
The point descriptor field succeeds in transferring the target point to \textit{unseen}~SE(3) poses, as well as to \textit{unseen} instances within the same class.
}
\label{fig:energy-right}
\vspace{-15pt}
\end{figure}

\paragraph{Validation -- \Figure{energy-left} and \Figure{energy-right}}
To validate the effectiveness of our descriptor fields, let us consider the following energy field:
\begin{equation}
E(\coord|\demopointcloud,\pointcloud,\demopoint)=\| \pointndf(\democoord|\demopointcloud) - \pointndf(\coord|\pointcloud) \|
\label{eq:energy2d}
\end{equation}
with its minimizer
\begin{equation}
\bar \coord = \underset{\coord}{\text{argmin}} \quad E(\coord|\demopointcloud,\pointcloud,\demopoint).
\label{eq:energy2d}
\end{equation}%
%
%
%
As shown in \Figure{energy-left}, given a reference point cloud $\demopointcloud$ and a reference point $\demopoint$, the minimizer $\bar \coord$ of Eq.~\ref{eq:energy2d} transfers the location of the reference point $\demopoint$ to the test-time object~$\pointcloud$.
%
%
%
In \Fig{energy-right}, we plot this energy for a reference point on the handle of a reference mug across different mug poses \textit{and} instances. 
The colors in the plot reflect that high-energy regions are far from the handle, whereas the energy deceases at positions closer to the handle.
We subsequently find that the transferred point $\bar \coord$ at the minimum of this energy field correctly corresponds to points on the handles across the different mugs, irrespective of their configuration. 
This validates that $f$ may transfer across object instances and generalize across $\SE{3}$ configurations.

\subsection{Neural Pose Descriptor Fields}
\label{sect:posendfs}

\begin{figure*}[t]
\includegraphics[width=\linewidth]{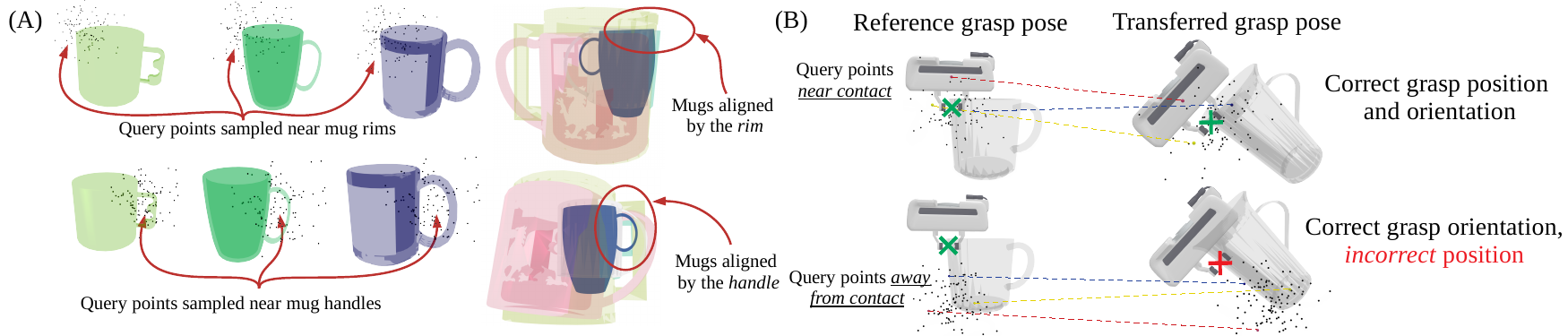}
\caption{\small \textbf{Effect of different query points} -- \textbf{(a)} (Top) Given a set of reference mugs and query points $\probes$ distributed near the rim of each mug, a set of differently sized test mugs can be aligned by their rim feature by finding a pose whose descriptor matches the average of the reference pose descriptors. (Bottom) Following this procedure with $\probes$ near the mug handles leads the same set of test mugs to be aligned by a different feature (the handle). This highlights the sensitivity to the location of $\probes$ when performing pose transfer. \textbf{(b)} This sensitivity has important implications when transferring gripper poses for grasping: (Top) When the points in $\probes$ are distributed near the rim of the mug and are used to transfer a grasp pose to a taller mug, the gripper position remains near the rim and the grasp can succeed. (Bottom) In contrast, placing query points near the bottom of the mug leads to a transferred pose that is biased toward the bottom of the taller mug, resulting in a grasp that will fail due to collision with the object.}
\label{fig:query-point-intuition}
\end{figure*}

\iftrue{
The previous section discussed how \ourshort{}s induce an energy that can be minimized for transferring points across object instances.
%
%
However, in manipulation tasks, we need to solve not only for a position (which may only denote, e.g., a single contact location) but also for the \emph{orientation} of an external rigid body such as the gripper.
For example, grasping the rim of a mug requires not only the correct contact position on the rim but also an orientation that enables the fingers to close around the inner and outer surface of the rim. If a grasp were attempted at the rim with an orientation that approached from the side of the mug, it wouldn't work. Similarly, to hang a mug on a rack by its handle, we must not only detect a point in the opening of the handle but also the orientation that allows the rack to pass through this opening.

%
Generally speaking, our demonstrations regularly consist of a point cloud $\demopointcloud$ along with a world-frame \emph{pose} $\demopose \in$ \sethree{} of some rigid body $\object$ in the vicinity of $\demopointcloud$ ($\object$ could be a gripper or a supporting object, like a rack or a shelf). We now wish to transfer both the position and orientation components of this pose when presented with a new point cloud.
In this section, we will leverage $\pointndf$ to find an equivalent pose of the rigid body $\object$ that reproduces the same task for a new object instance defined by its point cloud $\pointcloud$. 
%

We approach this from the perspective of defining a task-specific local coordinate frame, computing the pose $\demoposerel$ of external object $\object$ in this local frame, and solving for the corresponding local frame when presented with a new object instance.
After finding this corresponding frame, we use the same relative pose $\demoposerel$ in this detected frame to compute a new world-frame pose $\pose$ for object $\object$.
%
%
We leverage our knowledge about the pose $\demopose$ of object $\object$ to aid in parameterizing the pose of the local frame by \emph{fixing} the relative pose $\demoposerel$ to be the identity matrix $\mathbf{I}^{4}$, i.e. we constrain the local frame specified in the demonstrations to \emph{exactly align with the body frame defining the pose $\demopose$ in the world}. The result is that we can directly parameterize the resulting pose $\pose$ by the pose of the detected local frame for the new instance.

With this setup, an initial decision is how to encode local reference frames expressed as \sethree{} poses.
Our approach is guided by the observation that we can attach a reference frame to three or more (non-collinear) points which are constrained to move together rigidly, and establish a one-to-one mapping between these points and the configuration of the reference frame. Therefore, by initializing such a set of \emph{query points} $\probes \in \mathbb{R}^{3\times N}$ in a known canonical configuration, we can represent a local frame represented by an \sethree{} transformation $\pose$ via the action of $\pose$ on $\probes$.
%
%
%
%
$\pose$ is then represented via the coordinates of the \emph{transformed query point cloud} $\pose \probes_h$ (where $\probes_h$ denotes $\probes$ expressed with homogeneous coordinates). 

We now define a \emph{Neural Pose Descriptor Field} as the concatenated point descriptors of the individual points in $\pose\probes_h$:
\begin{equation}
\posedescriptor = \posendf(\pose | \pointcloud) = \bigoplus_{\coord_i \in \probes_h} \pointndf(\pose \coord_i |\pointcloud)
\label{eq:pose_enc}
\end{equation}
$\posendf$ maps a point cloud $\pointcloud$ and an \sethree{} transformation $\pose$ to a category-level pose descriptor, which we call $\posedescriptor$. \Fig{overview-right} shows a visualization of the architecture of $\posendf$.
Note that $\posendf$ inherits  \sethree{}-equivariance from $\pointndf$, and is thus similarly guaranteed to generalize across all 6-DoF object configurations of $\pointcloud$.
%

Similar to transferring individual points by minimizing point descriptor distances (\Fig{energy-left}), this encoding enables us to transfer a local frame with a reference pose $\demopose$ when provided with a new point cloud by finding the pose $\pose$ of the query point set $\probes$ that minimizes the distance to the descriptor $\hat{\posedescriptor} = \posendf(\demopose | \demopointcloud)$ (our approach for performing this minimization is described at the end of this sub-section). 
However, an important remaining decision is the choice of points $\coord_i \in \probes$. Any set of three or more points is equally sufficient to represent a reference pose, but the \emph{position of these points relative to} $\demopointcloud$ has a significant impact on what solutions are obtained when performing pose transfer. In particular, since we represent poses as the concatenation of individual point descriptors, the location of each $\coord_i$ in the demonstration fundamentally determines \emph{which features of the object we are aligning the rigid body to}. For instance, placing $\coord_i$ in the vicinity of the \emph{handle} of a mug would lead to a pose descriptor sensitive to the position of the \emph{handle} across mug instances. \Fig{query-point-intuition} highlights this issue by visualizing the effect of different ways of distributing the points in $\probes$. To select a set of points that is in the vicinity of the contact that occurs with object $\object$, we find that a robust heuristic is to sample points uniformly at random from within the bounding box of the rigid body $\object$.
}\fi

\begin{figure}[t]
\vspace{-10pt}
\includegraphics[width=\linewidth]{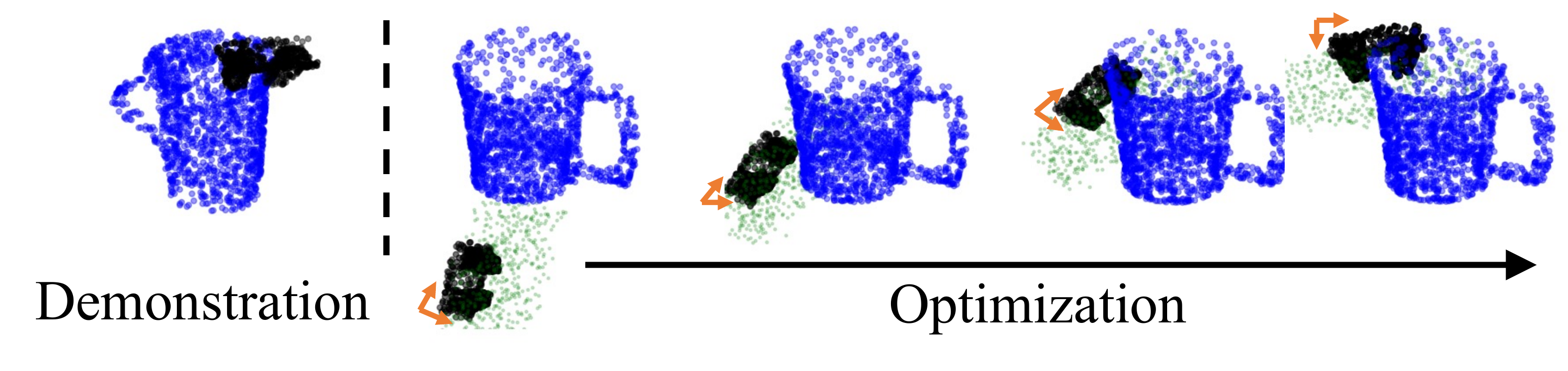}
\vspace{-15pt}
\caption{\small \textbf{Pose regression with \ourshort{}s} --
Given a demonstration point cloud and gripper pose (left), our method enables solving for the gripper pose (orange) for grasping an unseen object instance (right, blue) by minimizing the difference between demonstration and test pose descriptors, defined via the gripper query point cloud (green).}
\label{fig:optimization}
\vspace{-5pt}
\end{figure}

\paragraph{Pose regression with \ourshort{}s}
Similar to how $\pointndf$ induces an energy over \emph{coordinates} across object instances (see \Figure{energy-left} and \eq{energy2d}), $\posendf$ induces an energy over \emph{poses}.
We start with a tuple $(\demopose, \demopointcloud, \object)$ pairing pose $\demopose$ of rigid body $\object$ to a point cloud $\demopointcloud$. Then, given a novel object instance represented by its point cloud~$\pointcloud$, we can compute a pose $\pose$ such that the relative configuration between $\pointcloud$ and $\object$ at pose $\pose$ corresponds to the relative configuration between $\demopointcloud$ and $\object$ at pose $\demopose$. 
We initialize $\pose = (\mathbf{R}, \mathbf{t})$ at random and optimize the translation $\mathbf{t}$ and rotation $\mathbf{R}$ (parameterized via axis-angle) to minimize the L1 distance between the descriptors of~$\demopose$ and $\pose$:
%
\begin{align}
\bar \pose = \underset{\pose}{\text{argmin}} \|
 \posendf(\pose | \pointcloud) -
\posendf(\demopose | \demopointcloud)  \|
\label{eq:energy}
\end{align}
%
%
We solve this directly via iterative optimization (ADAM~\cite{kingma2014adam}), minimizing the distance between spatial descriptors of our target pose and our sought-after pose by back-propagating the norm of the differences through $\posedescriptor$. 
In \Fig{optimization} we visualize the optimization steps taken by~\eq{energy} for optimizing a grasp pose of the end-effector.
While we provide an in-depth evaluation in the experiments section, this result is representative in that the end-effector reliably and robustly converges to the correct orientation and location on the object.


\vspace{-3pt}
\subsection{Few-shot imitation learning with \ourshort{}s}
\label{sect:obj_manip}
We are now ready to use \ours{}s to acquire a pick-and-place skill for a category of objects from only a \textit{handful} of demonstrations.
%
For each category, we are provided with a set of $K$ demonstrations, $\{\mathcal{D}_{i}\}_{i=1}^{K}$.
Each demonstration $\mathcal{D}_i=(\pointcloud^{i}, \pose_{pick}^i, \pose_{rel}^i)$ is a tuple of a (potentially partial) point cloud of the object $\pointcloud^{i}$, and two poses: the end-effector pose before grasping, $\pose_{pick}^i$, and the relative pose $\pose^{i}_{rel}$ that transforms the grasp pose to the place pose via $\pose^{i}_{place} = \pose^{i}_{rel}\pose^{i}_{pick}$. 
%
%
First, we obtain $\probes_{pick}$ and $\probes_{place}$ to represent the gripper and placement surface, respectively.
%
%
%
We then leverage~\eq{pose_enc} to encode each pose $\pose^i_{*}$ into its vector of descriptors $\mathcal{Z}^i_{*}$, conditional on the respective object point cloud $\pointcloud^i$, obtaining a set of spatial descriptor tuples $\{(\mathcal{Z}^i_{pick}, \mathcal{Z}^i_{rel})\}_{i=1}^{K}$. 
%
%
Finally, this set of descriptors is averaged over the $K$ demonstrations to obtain \emph{single} pick and place descriptors $\bar{\mathcal{Z}}_{pick}$ and $\bar{\mathcal{Z}}_{rel}$. 
When a new object is placed in the scene at test time, we obtain a point cloud $\pointcloud^{test}$ and leverage~\eq{energy} to recover $\pose^{test}_{pick}$ and $\pose^{test}_{rel}$ by minimizing the distance to spatial descriptors $\bar{\mathcal{Z}}_{pick}$ and $\bar{\mathcal{Z}}_{rel}$. 
%
We rely on off-the-shelf inverse kinematics and motion planning algorithms to execute the final predicted pick-and-place task. 
\section{Experiments: Design and Setup}
Our experiments are designed to evaluate how effective our method is at generalizing pick-and-place tasks from a small number of demonstrations. 
In particular, we seek to answer three key questions: (1)~How well do \ourshort{}s enable manipulation of unseen objects in unseen poses? (2)~What impact does the parameterization of \ourshort{}s have on its performance? (3)~Can \ourshort{}s transfer to a real robot?

\begin{figure*}[t]
\includegraphics{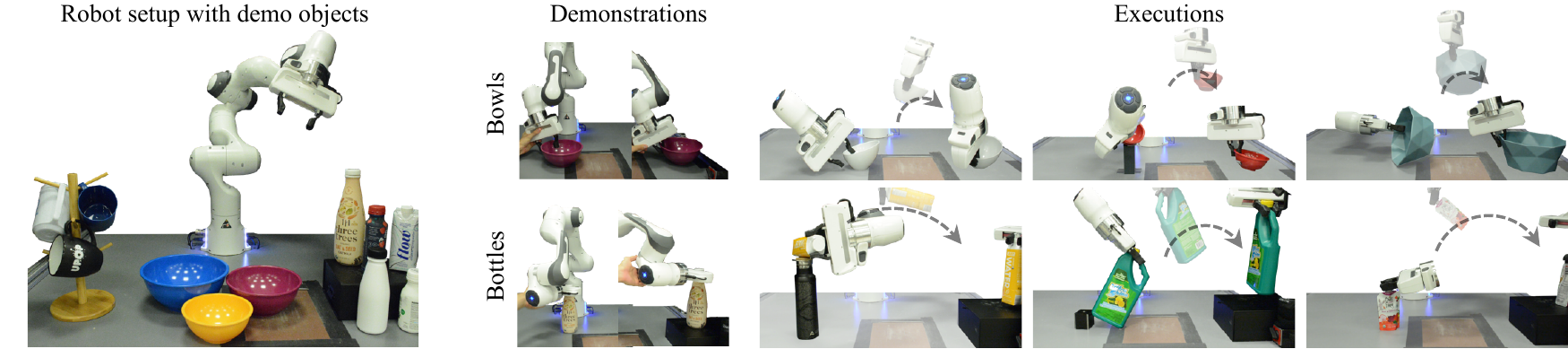}
\caption{\small Example executions of \ourshort on real bottles and bowls.
Ten demonstrations with an object in an upright pose were used per category. \ourshort{}s enable inferring 6-DoF poses for both picking and placing \emph{unseen} object instances in \emph{out-of-distribution} poses.} %
\label{fig:experiments}
\vspace{-15pt}
\end{figure*}

\paragraph{Robot Environment Setup}
Our environment includes a Franka Panda arm on 
a table with a depth camera at each table corner. The depth cameras are extrinsically calibrated to obtain fused point clouds expressed in the robot's base frame. For our quantitative experiments we simulate the environment in PyBullet~\cite{coumans2016pybullet}. Depending on the task, an additional object such as a rack or a shelf is mounted somewhere on the table to act as a placement/hanging surface; see~\Figure{experiments}.

\paragraph{Task Setup} We provide 10 demonstrations for each task, and measure execution success rates on unseen object instances with randomly sampled initial poses and a random uniform scaling applied.
We assume a segmented object point cloud and a static environment that remains fixed between demonstration-time and test-time.
We consider three separate tasks in both the simulated and real environment: 1)~grasping a mug by the rim and hanging it on a rack by the handle 2)~grasping a bowl by the rim and placing it upright on a shelf 3)~grasping the top of a bottle from the side and placing it upright on a shelf. 
In simulation, we utilize ShapeNet~\cite{chang2015shapenet} meshes for each object class, where we filter out a subset of meshes that are incompatible with the tasks.
%

\paragraph{Baselines}
We run a detailed quantitative comparison with a pick-and-place pipeline utilizing Dense Object Nets~(\don)~\cite{florence2018dense}. Our pipeline detects grasp poses following~\cite{florence2018dense} using demonstrated grasp points. To infer object placement, we label a set of semantic keypoints in demonstrations and utilize the \don correspondence model with depth to obtain corresponding 3D keypoints on test objects. We then estimate the relative transformation for placing by optimally registering the detected points to the final configuration of the corresponding points from the demonstrations using SVD.  
%

We also attempted to benchmark with recently proposed TransporterNets~\cite{zeng2020transporter}. However, the model in \cite{zeng2020transporter} is primarily applied to planar tasks that only require top-down pick-and-place. While we were able to reproduce the impressive capabilities of their model in the subset of our tasks in which top-down grasping is sufficient (92\% success rate at grasping the rim of a mug), several attempts at implementing a 6-DoF extended version that predicts the remaining rotational and $z$-height degrees of freedom (for grasping and placing) failed to achieve success rate above 10\%. 

\paragraph{Evaluation Metrics}
To quantify the capabilities of each method, we measure success rates for grasping (stable object contact after grasp close) and placing (stable contact with placement surface), along with overall success, corresponding to both grasp and placement success. 

\paragraph{Training Details}
To pretrain \don ~\cite{florence2018dense} and \ourshort{}, we generate a dataset of 100,000 objects of mug, bowl and bottle categories at random tabletop poses. For each object, 300 RGB-D views with labeled dense correspondences are used to train \don, while we train \ourshort{} with point clouds captured from four static depth cameras. 
RGB-D images of the objects are rendered with PyBullet. While \don requires separate models for shapes in each category, we train a single instance of \ourshort{} on shapes across all categories. We train \ourshort using an occupancy network $\Phi(\coord,\encoder(\pointcloud))$ to reconstruct 3D shapes given the captured depth maps and train \don utilizing the author's provided codebase.

\section{Experiments: Results}
\label{sec: experiments}

\begin{table}
\footnotesize
\setlength{\tabcolsep}{2.3pt}
\centering
\resizebox{\linewidth}{!}{
\begin{tabular}{@{}lccccccccc@{}}
    \toprule
      & \multicolumn{3}{c}{Mug} & \multicolumn{3}{c}{Bowl} & \multicolumn{3}{c}{Bottle}  \\
     \cmidrule(lr){2-4} \cmidrule(lr){5-7} \cmidrule(lr){8-10} 
      &  Grasp & Place & Overall &  Grasp & Place & Overall &  Grasp & Place & Overall \\
     \midrule
      \textbf{Upright Pose}\\
      \quad\quad\don ~\cite{florence2018dense} & 0.91 & 0.50 & 0.45  & 0.50 &  0.35 & 0.11  & 0.79 & 0.24 & 0.24\\
      \quad\quad\ourshort & \textbf{0.96} & \textbf{0.92} & \textbf{0.88} & \textbf{0.91} & \textbf{1.00} & \textbf{0.91}  & \textbf{0.87} & \textbf{1.00} & \textbf{0.87} \\
      \midrule
      \textbf{Arbitrary Pose}\\
      \quad\quad\don ~\cite{florence2018dense} & 0.35 & 0.45 & 0.17 & 0.08 & 0.20 & 0.00 & 0.05 & 0.02 & 0.01 \\
      \quad\quad\ourshort & \textbf{0.78} & \textbf{0.75} & \textbf{0.58} & \textbf{0.79} & \textbf{0.97} & \textbf{0.78} & \textbf{0.78} & \textbf{0.99} & \textbf{0.77} \\
    \bottomrule
    
\end{tabular}
    
}
\caption{\small \textbf{Unseen instance pick-and-place success rates in simulation.} For objects in upright poses (top row), \ourshort{}s perform on par with \don{} on grasp success rate, but outperforms \don{} on overall pick-and-place success rate. For objects in arbitrary poses (bottom row), \don{}'s performance suffers, while \ourshort{}s maintains higher success rates due to their equivariance to \sethree{} transformations.}
\label{tbl:benchmark}
\vspace{-10pt}
\end{table}
\begin{figure}[t]
\includegraphics[width=\linewidth]{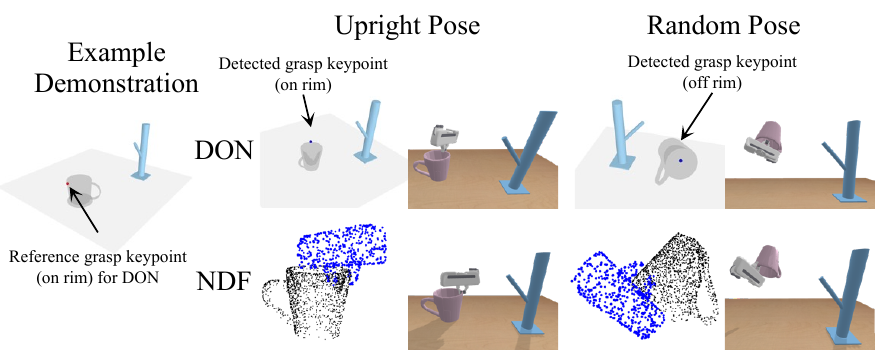}
\vspace{-15pt}
\caption{\small \textbf{Qualitative Examples of Grasp Predictions} -- Both \don and \ourshort predict successful grasps on upright mugs. When mugs exhibit arbitrary poses, \don fails to detect the correct keypoint for grasping, while our method still successfully infers a grasp.}
\label{fig:baseline-compare}
\vspace{-20pt}
\end{figure}

We conduct experiments in simulation to compare the performance of \ourshort{}s with Dense Object Nets (\don)~\cite{florence2018dense} on three different object classes, and different pose configurations of each object.
We then conduct ablation studies of the choice of parameterizing \ourshort{}s as the concatenation of pretrained occupancy network activations, as well as the effect of the number of demonstrations.
Finally, we apply our full model to a real robot, and validate that the proposed method generalizes to out-of-distribution object configurations. 

\subsection{Simulation Experiments}

\paragraph{Upright Pose} First, we consider the ability to transfer manipulation skills to novel objects in different upright poses. We find that across mugs, bowls, and bottles, \ourshort{}s dramatically outperform \don{} on placing, and perform significantly better on grasping~(\Table{benchmark}, top). We find that \don{}'s failures are usually a function of either insufficient precision in keypoint predictions, or failed registration of test-time keypoints to the demonstration keypoints. We find that even if predicted keypoints locations are semantically correct, the place may still fail when the \emph{relative locations} of keypoints to each other are too different from the demonstration objects. This may happen, for instance, if the object is significantly smaller, or the shape is otherwise significantly different. In contrast, the proposed method matches descriptors in a learned, highly over-parameterized latent space, and is significantly more robust in solving for placement poses.

\paragraph{Arbitrary Pose}
Next, we consider a harder setting: while the demonstrations are all performed on upright-posed objects, the robot must subsequently execute the task on objects in \emph{arbitrary} \sethree{} poses. 
In this setting, we find that the performance of \don{} suffers significantly, even though we trained \don{} on a large dataset of images of objects in different poses.
In contrast, we find that \ourshort{}'s performance, while not at the same level as in the upright task, suffers dramatically less, maintaining a high pick-and-place success rate~(\Table{benchmark}, bottom).
\Figure{baseline-compare} highlights an example to illustrate this performance gap.
The drop in our method's performance can be attributed to the fact that while provably equivariant to rotations and translations, the PointNet encoder is not perfectly robust to unobserved occlusions and disocclusions of the object point cloud: pointclouds might be missing parts previously observed, or contain parts that were previously unobserved. For instance, if only upright mugs were observed, the encoder has not previously seen the bottom of a mug.

\subsection{Analysis}
We now analyze \ourshort{}'s dependence on the occupancy network parameterization, the number of demonstrations, and the size of the query point cloud used for encoding pose descriptors. We run our analysis on the upright mug task. 

\begin{table}[t]
\footnotesize
\setlength{\tabcolsep}{2.3pt}
\centering
\resizebox{\linewidth}{!}{
\begin{tabular}{@{}cccccccccccc@{}}
    \toprule
     \multicolumn{3}{c}{Random \ourshort{}}& \multicolumn{3}{c}{Last Layer OccNet} & \multicolumn{3}{c}{First Layer OccNet} & \multicolumn{3}{c}{All Layer OccNet} \\
     \cmidrule(lr){1-3} \cmidrule(lr){4-6} \cmidrule(lr){7-9} \cmidrule(lr){10-12}
     Grasp & Place & Overall & Grasp & Place & Overall  & Grasp & Place & Overall & Grasp & Place & Overall \\
     \midrule
     0.42 & 0.00 & 0.00 & 0.75 & 0.88 & 0.65 & 0.77 & 0.84 & 0.65 & \textbf{0.96} & \textbf{0.92} & \textbf{0.88}\\
    \bottomrule
\end{tabular}
} 
\caption{\small Success rates of \ourshort{}s with different descriptors.}
\label{tbl:ablation}
\end{table}

\begin{table}[t]
\footnotesize
\setlength{\tabcolsep}{2.3pt}
\centering
\resizebox{\linewidth}{!}{
\begin{tabular}{@{}ccccccccccccccc@{}}
    \toprule
     \multicolumn{3}{c}{0.1}& \multicolumn{3}{c}{0.5} & \multicolumn{3}{c}{1.0} & \multicolumn{3}{c}{2.0} & \multicolumn{3}{c}{10.0}\\
     \cmidrule(lr){1-3} \cmidrule(lr){4-6} \cmidrule(lr){7-9} \cmidrule(lr){10-12} \cmidrule(lr){13-15}
     G & P & O & G & P & O  & G & P & O & G & P & O & G & P & O \\
     \midrule
     0.77 & 0.53 & 0.39 & 0.77 & 0.85 & 0.63 & \textbf{0.96} & \textbf{0.92} & \textbf{0.88} & 0.78 & 0.56 & 0.42 & 0.73 & 0.39 & 0.27\\
    \bottomrule
\end{tabular}
} 
\caption{\small Effect of representing pose descriptors with differently scaled query point clouds $\probes$.}
\label{tbl:query-pt-scale}
\vspace{-10pt}
\end{table}

\begin{figure}[t]
\centering
\begin{minipage}[c]{.55\linewidth}
\footnotesize
\setlength{\tabcolsep}{5.5pt}
\centering
\vspace{-5pt}
\begin{tabular}{cccc}
    \toprule
    Model & 1  & 5  & 10 \\
     \midrule
     \don ~\cite{florence2018dense} & 0.32 & 0.36 & 0.45\\
     \ourshort & \textbf{0.46} & \textbf{0.70} & \textbf{0.88}\\
    \bottomrule
\end{tabular}
\end{minipage}
\hfill
\begin{minipage}[c]{.40\linewidth}
\captionof{table}{\small Overall success of \ourshort{}s and \don as a function of total example demonstrations.}
\label{tbl:no_demonstration}
\end{minipage}
\vspace{-15pt}
\end{figure}
\paragraph{Neural Descriptors}
Full \ourshort{}s are parameterized as the concatenation of the activations of all layers of an occupancy network trained for 3D reconstruction. In \Table{ablation}, we analyze the effect of parameterizing \ourshort{}s with features from a randomly initialized occupancy network, as well as with only the first- or last-layer activations of a trained occupancy network.
We find that utilizing all activations obtains the best performance by a large margin. This validates our assumptions on occupancy networks as a hierarchical feature extractor, and the task of 3D reconstruction as an important part of learning informative features.

\paragraph{Query Point Cloud Scaling}
We further study the effect of the scale of the query point cloud $\probes$ for representing the grasping and placing pose descriptors. 
In \Table{query-pt-scale} we show that our choice of sampling in the bounding box of the rigid body that interacts with the object is a robust heuristic, while scaling $\probes$ up or down reduces the performance.

\paragraph{Number of Demonstrations}
We also analyze the impact of demonstration number on the performance of \ourshort{}s and \don{} on the upright mug pick-and-place task. Please see \Table{no_demonstration} for quantitative results. We find that while the performance of \ourshort{}s decreases significantly in the single-demonstration case, it still significantly outperforms \don{}, and more demonstrations yield significant performance gains.  

%

\subsection{Real World Execution}
Finally, we validate that \ourshort{}s enable manipulation of novel object instances in novel poses on the real robot. We record ten pick-and-place demonstrations on mugs, bowls, bottles in \emph{upright poses}. We then execute the same pick-and-place task on novel instances of real mugs, bowls, and bottles in a variety of different, often challenging, configurations. Please see \Figure{experiments} both for a visualization of the demonstrations and the qualitative results, as well as the \textbf{supplementary video} for sample videos of each of the real world task executions.
\section{Related Work}
\label{sec: related}

\subsection{Generalizable Manipulation}
Our work builds upon a rich line of research on imitation learning for manipulation.
For known objects, one may rely on pose estimation~\cite{yoon2003real, zhu2014single, Schulman2016}, however, this does not enable category-level manipulation.
Template-matching with coarse 3D primitives~\cite{miller2003automatic, harada2013probabilistic, LIS277_shapeBasedSkillTransfer_icra2021} or non-rigid registration~\cite{Schulman2016} can enable generalization across changes in shape and pose, but suffers when objects deviate significantly from the primitive or test and reference scene are too different.
Direct learning of pick-and-place policies meanwhile requires large amounts of data from demonstrations~\cite{berscheid2020self,gualtieri2018pick,song2020grasping}.

Our work is closely related to recent work leveraging category-level keypoints as an object representation for transferrable robotic manipulation.
Keypoints can either be predicted directly~\cite{gao2019kpam,manuelli2019kpam, gao2021kpam}, requiring a large, human-annotated dataset, or can be chosen among a set of self-supervised category-level object correspondences~\cite{florence2018dense,sundaresan2020learning}.
However, keypoints must be carefully chosen to properly constrain manipulation poses, with outcomes sensitive to both keypoint choice and accuracy.
%
Both approaches use 2D convolutional neural networks (CNNs) for prediction. As 2D CNNs are only equivariant to shifts of the object parallel to the image plane, these methods require observing images of objects from \emph{all possible} rotations and translations at training time, and even then do not guarantee that keypoints are consistent across 6-DoF configurations of object instances.
Transporter Nets~\cite{zeng2020transporter} predict manipulation poses via a CNN over orthographic, top-down views, equipping the model with equivariance to in-plane, 2D translations of objects. However, this approach struggles to predict arbitrary 6-DoF poses, and is not equivariant to full 3D rotations and translations.

\ours{}s enable transferring observed manipulation poses across an object category using task-agnostic, self-supervised pre-training, without human-labeled keypoints, and are fully equivariant to \sethree{} transformations. We demonstrate imitation of full pick-and-place tasks for unseen object configurations from a small handful of demonstrations, and significantly outperform baselines based on correspondence predicted in 2D.

\subsection{Neural Fields and Neural Scene Representations} 
Our approach leverages neural implicit representations to parameterize a continuous descriptor field which represents a manipulated object. Most saliently, such fields have been proposed to represent 3D geometry ~\cite{Niemeyer2019ICCV,chen2019learning,park2019deepsdf,peng2020convolutional,rebain2021deep}, appearance \cite{sitzmann2019srns,mildenhall2020nerf,Niemeyer2020DVR,yariv2020multiview,sitzmann2021lfns,saito2019pifu,yu2020pixelnerf}, and tactile properties~\cite{gao2021objectfolder}. They offer several benefits over conventional discrete representations: due to their continuous nature, they parameterize scene surfaces with ``infinite resolution''. Furthermore, their functional nature enables the principled incorporation of symmetries, such as \sothree{} equivariance~\cite{deng2021vector,zhu2021correspondence}. Their functional nature further enables the construction of latent spaces that encode class information as well as 3D correspondence~\cite{kohli2020semantic,deng2021deformed,sitzmann2020metasdf}.
Lastly, neural fields have been leveraged to find unknown camera poses in 3D reconstruction tasks~\cite{yen2020inerf,lin2021barf}.

\section{Discussion and Conclusion}
\label{sec:conclusion}

Several limitations and avenues for future work remain.
While this approach is in principle applicable to non-rigid objects, this remains to be tested, and extensions based on recent work on non-rigid scenes in 3D reconstruction and novel view synthesis~\cite{niemeyer2019occupancy,li2020neural, pumarola2020dnerf,  park2020nerfies, xian2021space, du2021nerflow} might be necessary.
Further, \ourshort{}s only define transferable energy landscapes over \emph{poses} and \emph{points}: future work may explore integrating such energy functions with trajectory optimization to enable \ourshort{}s to transfer to full trajectories.
Furthermore, we assume the placement target remains static: future work may explore similarly inferring an object-centric representation of the placement target. 

In summary, this work introduces \ours{}s as object representations that allow few-shot imitation learning of manipulation tasks, with only task-agnostic pre-training in the form of 3D geometry reconstruction, and without the need for further training at imitation learning time.
We build on prior work using dense descriptors for robotics, neural fields, and geometric machine learning to develop dense descriptors that both generalize across instances and provably generalize across \sethree{} configurations, which we show enables our approach to apply to novel objects in both novel rotations and translations, where 2D dense descriptors are insufficient.

\section{Acknowledgement}
This work is supported by DARPA under CW3031624 (Transfer, Augmentation and Automatic Learning with Less Labels) and the Machine Common Sense program, Singapore DSTA under DST00OECI20300823 (New Representations for Vision), Amazon Research Awards, and the NSF AI Institute for Artificial Intelligence and Fundamental Interactions (IAIFI). Anthony and Yilun are supported in part by NSF Graduate Research Fellowships. We would like to thank Leslie Kaelbling and Tomas Lozano-Perez for helpful discussions.
\newpage




\renewcommand*{\bibfont}{\footnotesize}
\begin{flushright}
\printbibliography 

@inproceedings{florence2018dense,
  title={Dense Object Nets: Learning Dense Visual Object Descriptors By and For Robotic Manipulation},
  author={Florence, Peter R and Manuelli, Lucas and Tedrake, Russ},
  booktitle={Conference on Robot Learning},
  pages={373--385},
  year={2018},
  organization={PMLR}
}

@article{sundaresan2020learning,
  title={Learning Rope Manipulation Policies Using Dense Object Descriptors Trained on Synthetic Depth Data},
  author={Sundaresan, Priya and Grannen, Jennifer and Thananjeyan, Brijen and Balakrishna, Ashwin and Laskey, Michael and Stone, Kevin and Gonzalez, Joseph E and Goldberg, Ken},
  journal={arXiv preprint arXiv:2003.01835},
  year={2020}
}

@article{manuelli2019kpam,
  title={kpam: Keypoint affordances for category-level robotic manipulation},
  author={Manuelli, Lucas and Gao, Wei and Florence, Peter and Tedrake, Russ},
  journal={arXiv preprint arXiv:1903.06684},
  year={2019}
}

@article{gao2019kpam,
  title={kPAM-SC: Generalizable Manipulation Planning using KeyPoint Affordance and Shape Completion},
  author={Gao, Wei and Tedrake, Russ},
  journal={arXiv preprint arXiv:1909.06980},
  year={2019}
}

@inproceedings{qi2017pointnet++,
  title={Pointnet++: Deep hierarchical feature learning on point sets in a metric space},
  author={Qi, Charles Ruizhongtai and Yi, Li and Su, Hao and Guibas, Leonidas J},
  booktitle={Advances in neural information processing systems},
  pages={5099--5108},
  year={2017}
}

@inproceedings{gualtieri2018pick,
  title={Pick and place without geometric object models},
  author={Gualtieri, Marcus and ten Pas, Andreas and Platt, Robert},
  booktitle={2018 IEEE International Conference on Robotics and Automation  },
  pages={7433--7440},
  year={2018},
  organization={IEEE}
}

@article{coumans2016pybullet,
  title={Pybullet, a python module for physics simulation for games, robotics and machine learning},
  author={Coumans, Erwin and Bai, Yunfei},
  journal={GitHub repository},
  year={2016}
}

@article{kingma2014adam,
  title={Adam: A method for stochastic optimization},
  author={Kingma, Diederik P and Ba, Jimmy},
  journal={arXiv preprint arXiv:1412.6980},
  year={2014}
}

@article{zeng2020transporter,
    title={Transporter Networks: Rearranging the Visual World for Robotic Manipulation},
    author={Zeng, Andy and Florence, Pete and Tompson, Jonathan and Welker, Stefan and Chien, Jonathan and Attarian, Maria and Armstrong, Travis and Krasin, Ivan and Duong, Dan and Sindhwani, Vikas and Lee, Johnny},
    journal={Conference on Robot Learning (CoRL)},
    year={2020}
}

@InProceedings{LIS277_shapeBasedSkillTransfer_icra2021,
title = {Shape-Based Transfer of Generic Skills},
author = {Skye Thompson and Leslie Pack Kaelbling and Tomas Lozano-Perez},
year = {2021},
url = {https://lis.csail.mit.edu/wp-content/uploads/2021/05/thompson_icra_2021_compressed.pdf},
booktitle = {Proc. of The International Conference in Robotics and Automation (ICRA)},
}

@Inbook{Schulman2016,
author="Schulman, John
and Ho, Jonathan
and Lee, Cameron
and Abbeel, Pieter",
editor="Inaba, Masayuki
and Corke, Peter",
title="Learning from Demonstrations Through the Use of Non-rigid Registration",
bookTitle="Robotics Research: The 16th International Symposium ISRR",
year="2016",
publisher="Springer International Publishing",
address="Cham",
pages="339--354",
isbn="978-3-319-28872-7",
doi="10.1007/978-3-319-28872-7_20",
url="https://doi.org/10.1007/978-3-319-28872-7_20"
}

@inproceedings{yen2020inerf,
  title={{iNeRF}: Inverting Neural Radiance Fields for Pose Estimation},
  author={Lin Yen-Chen and Pete Florence and Jonathan T. Barron and Alberto Rodriguez and Phillip Isola and Tsung-Yi Lin},
  booktitle={IEEE/RSJ International Conference on Intelligent Robots and Systems ({IROS})},
  year={2021}
}

@inproceedings{kohli2020semantic,
  title={Semantic implicit neural scene representations with semi-supervised training},
  author={Kohli, Amit Pal Singh and Sitzmann, Vincent and Wetzstein, Gordon},
  booktitle={2020 International Conference on 3D Vision (3DV)},
  pages={423--433},
  year={2020},
  organization={IEEE}
}

@inproceedings{lin2021barf,
  title={BARF: Bundle-Adjusting Neural Radiance Fields},
  author={Lin, Chen-Hsuan and Ma, Wei-Chiu and Torralba, Antonio and Lucey, Simon},
  booktitle={IEEE International Conference on Computer Vision ({ICCV})},
  year={2021}
}

@article{gao2021kpam,
  title={kPAM 2.0: Feedback Control for Category-Level Robotic Manipulation},
  author={Gao, Wei and Tedrake, Russ},
  journal={IEEE Robotics and Automation Letters},
  volume={6},
  number={2},
  pages={2962--2969},
  year={2021},
  publisher={IEEE}
}

@article{argall2009survey,
  title={A survey of robot learning from demonstration},
  author={Argall, Brenna D and Chernova, Sonia and Veloso, Manuela and Browning, Brett},
  journal={Robotics and autonomous systems},
  year={2009}
}

@article{schaal1999imitation,
  title={Is imitation learning the route to humanoid robots?},
  author={Schaal, Stefan},
  journal={Trends in cognitive sciences},
  year={1999}
}

@inproceedings{pomerleau1989alvinn,
  title={{ALVINN}: An autonomous land vehicle in a neural network},
  author={Pomerleau, Dean A},
  booktitle={NIPS},
  year={1989}
}

@inproceedings{mescheder2019occupancy,
	title={Occupancy Networks: Learning 3D Reconstruction in Function Space},
	author={Mescheder, Lars and Oechsle, Michael and Niemeyer, Michael and Nowozin, Sebastian and Geiger, Andreas},
	booktitle={Proc. CVPR},
	year={2019}
}

@inproceedings{park2019deepsdf,
	title={DeepSDF: Learning Continuous Signed Distance Functions for Shape Representation},
	author={Park, Jeong Joon and Florence, Peter and Straub, Julian and Newcombe, Richard and Lovegrove, Steven},
	booktitle={Proc. CVPR},
	year={2019}
}

@inproceedings{sitzmann2019srns,
	author = {Sitzmann, Vincent
	and Zollh{\"o}fer, Michael
	and Wetzstein, Gordon},
	title = {Scene Representation Networks: Continuous 3D-Structure-Aware Neural Scene Representations},
	booktitle = {Proc. NeurIPS 2019},
	year={2019}
}

@inproceedings{mildenhall2020nerf,
	title={NeRF: Representing Scenes as Neural Radiance Fields for View Synthesis},
	author={Mildenhall, Ben and Srinivasan, Pratul P and Tancik, Matthew and Barron, Jonathan T and Ramamoorthi, Ravi and Ng, Ren},
	booktitle={Proc. ECCV},
	year={2020}
}

@inproceedings{peng2020convolutional,
	title={Convolutional occupancy networks},
	author={Peng, Songyou and Niemeyer, Michael and Mescheder, Lars and Pollefeys, Marc and Geiger, Andreas},
	booktitle={Proc. ECCV},
	year={2020}
}

@inproceedings{chen2019learning,
	title={Learning implicit fields for generative shape modeling},
	author={Chen, Zhiqin and Zhang, Hao},
	booktitle={Proc. CVPR},
	pages={5939--5948},
	year={2019}
}

@inproceedings{saito2019pifu,
	title={Pifu: Pixel-aligned implicit function for high-resolution clothed human digitization},
	author={Saito, Shunsuke and Huang, Zeng and Natsume, Ryota and Morishima, Shigeo and Kanazawa, Angjoo and Li, Hao},
	booktitle={Proc. ICCV},
	pages={2304--2314},
	year={2019}
}

@INPROCEEDINGS{Niemeyer2020DVR,
	author = {Michael Niemeyer and Lars Mescheder and Michael Oechsle and Andreas Geiger},
	title = {Differentiable Volumetric Rendering: Learning Implicit 3D Representations without 3D Supervision},
	booktitle = {Proc. CVPR},
	year = {2020}
}

@INPROCEEDINGS{Niemeyer2019ICCV,
	author = {Michael Niemeyer and Lars Mescheder and Michael Oechsle and Andreas Geiger},
	title = {Occupancy Flow: 4D Reconstruction by Learning Particle Dynamics},
	booktitle = {Proc. ICCV},
	year = {2019}
}

@inproceedings{chen2021learning,
  title={Learning continuous image representation with local implicit image function},
  author={Chen, Yinbo and Liu, Sifei and Wang, Xiaolong},
  booktitle={Proceedings of the IEEE/CVF Conference on Computer Vision and Pattern Recognition},
  pages={8628--8638},
  year={2021}
}

@article{sitzmann2020metasdf,
  title={Metasdf: Meta-learning signed distance functions},
  author={Sitzmann, Vincent and Chan, Eric R and Tucker, Richard and Snavely, Noah and Wetzstein, Gordon},
  journal={Proc. NeurIPS},
  year={2020}
}

@article{yariv2020multiview,
  title={Multiview neural surface reconstruction by disentangling geometry and appearance},
  author={Yariv, Lior and Kasten, Yoni and Moran, Dror and Galun, Meirav and Atzmon, Matan and Ronen, Basri and Lipman, Yaron},
  journal={Proc. NeurIPS},
  year={2020}
}

@article{yu2020pixelnerf,
  title={pixelNeRF: Neural Radiance Fields from One or Few Images},
  author={Yu, Alex and Ye, Vickie and Tancik, Matthew and Kanazawa, Angjoo},
  journal={Proc. CVPR},
  year={2020}
}

@article{chang2015shapenet,
	title={Shapenet: An information-rich 3d model repository},
	author={Chang, Angel X and Funkhouser, Thomas and Guibas, Leonidas and Hanrahan, Pat and Huang, Qixing and Li, Zimo and Savarese, Silvio and Savva, Manolis and Song, Shuran and Su, Hao and others},
	journal={arXiv preprint arXiv:1512.03012},
	year={2015}
}

@inproceedings{sitzmann2021lfns,
    author = {Sitzmann, Vincent
              and Rezchikov, Semon
              and Freeman, William T.
              and Tenenbaum, Joshua B.
              and Durand, Fredo},
    title = {Light Field Networks: Neural Scene Representations
             with Single-Evaluation Rendering},
    booktitle = {arXiv},
    year={2021}
}

@article{deng2021vector,
  title={Vector Neurons: A General Framework for SO (3)-Equivariant Networks},
  author={Deng, Congyue and Litany, Or and Duan, Yueqi and Poulenard, Adrien and Tagliasacchi, Andrea and Guibas, Leonidas},
  journal={arXiv preprint arXiv:2104.12229},
  year={2021}
}

@article{zhu2021correspondence,
  title={Correspondence-Free Point Cloud Registration with SO (3)-Equivariant Implicit Shape Representations},
  author={Zhu, Minghan and Ghaffari, Maani and Peng, Huei},
  journal={arXiv preprint arXiv:2107.10296},
  year={2021}
}

@inproceedings{deng2021deformed,
  title={Deformed implicit field: Modeling 3d shapes with learned dense correspondence},
  author={Deng, Yu and Yang, Jiaolong and Tong, Xin},
  booktitle={Proceedings of the IEEE/CVF Conference on Computer Vision and Pattern Recognition},
  pages={10286--10296},
  year={2021}
}

@inproceedings{niemeyer2019occupancy,
	title={Occupancy flow: 4d reconstruction by learning particle dynamics},
	author={Niemeyer, Michael and Mescheder, Lars and Oechsle, Michael and Geiger, Andreas},
	booktitle={Proceedings of the IEEE International Conference on Computer Vision},
	pages={5379--5389},
	year={2019}
}

@InProceedings{li2020neural,
  title={Neural Scene Flow Fields for Space-Time View Synthesis of Dynamic Scenes},
  author={Li, Zhengqi and Niklaus, Simon and Snavely, Noah and Wang, Oliver},
  booktitle = {Proceedings of the IEEE/CVF Conference on Computer Vision and Pattern Recognition (CVPR)},
  year={2021}
}

@article{pumarola2020dnerf,
  author={Albert Pumarola and Enric Corona and Gerard Pons-Moll and Francesc Moreno-Noguer},
  title={D-NeRF: Neural Radiance Fields for Dynamic Scenes},
  year={2020},
  journal={arXiv preprint arXiv:2011.13961},
}

@article{park2020nerfies,
  author    = {Park, Keunhong and Sinha, Utkarsh and Barron , Jonathan T. and Bouaziz, Sofien and Goldman, Dan B. 
               and Seitz, Steven M. 
               and Martin-Brualla, Ricardo},
  title     = {Deformable Neural Radiance Fields},
  journal   = {arXiv preprint arXiv:2011.12948},
  year      = {2020},
}

@inproceedings{xian2021space,
  author    = {Wenqi Xian and Jia-Bin Huang and Johannes Kopf and Changil Kim},
  title     = {Space-time Neural Irradiance Fields for Free-Viewpoint Video},
  booktitle = {Proceedings of the IEEE/CVF Conference on Computer Vision and Pattern Recognition (CVPR)},
  year      = {2021},
  pages     = {9421--9431},
}

@inproceedings{du2021nerflow,
  author    = {Yilun Du and Yinan Zhang and Hong-Xing Yu 
               and Joshua B. Tenenbaum and Jiajun Wu},
  title     = {Neural Radiance Flow for 4D View Synthesis and Video Processing},
  year      = {2021},
  booktitle   = {Proceedings of the IEEE/CVF International Conference
                 on Computer Vision},
}

@article{gao2021objectfolder,
  title={ObjectFolder: A Dataset of Objects with Implicit Visual, Auditory, and Tactile Representations},
  author={Gao, Ruohan and Chang, Yen-Yu and Mall, Shivani and Fei-Fei, Li and Wu, Jiajun},
  journal={arXiv preprint arXiv:2109.07991},
  year={2021}
}

@article{rebain2021deep,
  title={Deep Medial Fields},
  author={Rebain, Daniel and Li, Ke and Sitzmann, Vincent and Yazdani, Soroosh and Yi, Kwang Moo and Tagliasacchi, Andrea},
  journal={arXiv preprint arXiv:2106.03804},
  year={2021}
}

@inproceedings{yoon2003real,
  title={Real-time tracking and pose estimation for industrial objects using geometric features},
  author={Yoon, Youngrock and DeSouza, Guilherme N and Kak, Avinash C},
  booktitle={2003 IEEE International Conference on Robotics and Automation (Cat. No. 03CH37422)},
  volume={3},
  pages={3473--3478},
  year={2003},
  organization={IEEE}
}

@inproceedings{zhu2014single,
  title={Single image 3D object detection and pose estimation for grasping},
  author={Zhu, Menglong and Derpanis, Konstantinos G and Yang, Yinfei and Brahmbhatt, Samarth and Zhang, Mabel and Phillips, Cody and Lecce, Matthieu and Daniilidis, Kostas},
  booktitle={2014 IEEE International Conference on Robotics and Automation (ICRA)},
  pages={3936--3943},
  year={2014},
  organization={IEEE}
}

@article{berscheid2020self,
  title={Self-supervised learning for precise pick-and-place without object model},
  author={Berscheid, Lars and Mei{\ss}ner, Pascal and Kr{\"o}ger, Torsten},
  journal={IEEE Robotics and Automation Letters},
  volume={5},
  number={3},
  pages={4828--4835},
  year={2020},
  publisher={IEEE}
}

@inproceedings{miller2003automatic,
  title={Automatic grasp planning using shape primitives},
  author={Miller, Andrew T and Knoop, Steffen and Christensen, Henrik I and Allen, Peter K},
  booktitle={2003 IEEE International Conference on Robotics and Automation (Cat. No. 03CH37422)},
  volume={2},
  pages={1824--1829},
  year={2003},
  organization={IEEE}
}

@inproceedings{harada2013probabilistic,
  title={Probabilistic approach for object bin picking approximated by cylinders},
  author={Harada, Kensuke and Nagata, Kazuyuki and Tsuji, Tokuo and Yamanobe, Natsuki and Nakamura, Akira and Kawai, Yoshihiro},
  booktitle={2013 IEEE International Conference on Robotics and Automation},
  pages={3742--3747},
  year={2013},
  organization={IEEE}
}

@article{song2020grasping,
  title={Grasping in the wild: Learning 6dof closed-loop grasping from low-cost demonstrations},
  author={Song, Shuran and Zeng, Andy and Lee, Johnny and Funkhouser, Thomas},
  journal={IEEE Robotics and Automation Letters},
  volume={5},
  number={3},
  pages={4978--4985},
  year={2020},
  publisher={IEEE}
}
\end{flushright}

\end{document}